\documentclass[conference]{IEEEtran}
\IEEEoverridecommandlockouts
\usepackage{iccv}
\usepackage{times}
\usepackage{epsfig}
\usepackage{graphicx}
\usepackage{cite}
\usepackage{amssymb,amsfonts}
\usepackage{amsmath}
\usepackage{algorithmic}
\usepackage{textcomp}
\usepackage{xcolor}
\usepackage{float}
\def\BibTeX{{\rm B\kern-.05em{\sc i\kern-.025em b}\kern-.08em
    T\kern-.1667em\lower.7ex\hbox{E}\kern-.125emX}}

\iccvfinalcopy 

\def\iccvPaperID{12366} 

\ificcvfinal\fi
    
\makeatletter
\newcommand{\linebreakand}{%
  \end{@IEEEauthorhalign}
  \hfill\mbox{}\par
  \mbox{}\hfill\begin{@IEEEauthorhalign}
}

\makeatother
\begin{document}

\title{\ Atrous Space Bender U-Net (ASBU-Net/LogiNet)}

\author{Anurag Bansal ~~~~~~~~~~ Oleg Ostap ~~~~~~~~~~ Miguel Maestre Trueba ~~~~~~~~~~ Kristopher Perry\\ 
\IEEEauthorblockA{Logitech Inc.\\
{\tt\small\{anuragb, oostap, mmaestretrueba, kperry\}@logitech.com}
}
}

\maketitle
\ificcvfinal\thispagestyle{empty}\fi

\begin{abstract}
With recent advances in CNNs, exceptional improvements have been made in semantic segmentation of high resolution image in terms of accuracy and latency. However, challenges still remain in detecting objects in crowded scenes, large scale variations, partial occlusion, and distortions, while still maintaining mobility and latency. We introduce a fast and efficient convolutional neural network, ASBU-Net, for semantic segmentation of high resolution images that addresses these problems and uses no novelty layers for ease of quantization and embedded hardware support. ASBU-Net is based on a new feature extraction module, atrous space bender layer (ASBL), which is efficient in terms of computation and memory. The ASB layers form a building block that is used to make ASBNet. Since this network does not use any special layers it can be easily implemented, quantized and deployed on FPGAs and other hardware with limited memory. We present experiments on resource and accuracy trade-offs and show strong performance compared to other popular models.
\end{abstract}

\section{Introduction}
Ever since AlexNet \cite{b1} popularized deep convolutional neural networks by winning the ImageNet Challenge: ILSVRC 2012 \cite{b2}, everyone is trying to incorporate convolutional neural networks in their products as an end to end solution or a part of the state machine. The general research has been focused to improve accuracy of detection by making the networks deeper and more complicated \cite{b3}\cite{b4}\cite{b5}\cite{b6}\cite{b7}\cite{b8}. With the advances in the accuracy of detection, the need and demand for CNNs have been increasing everyday. In the recent years, there has been a slight shift in paradigm where focus moved towards having networks that can run on actual hardware with limited memory and compute capabilities \cite{b9}\cite{b10}\cite{b11}\cite{b12}\cite{b13}\cite{b14}\cite{b15}. Thus CNNs are no longer just running on cutting-edge GPUs in university research labs, but in most of the electronic devices being used for video collaboration, autonomous driving, warehouse automation, augmented reality, etc.

In recent years, visual data has grown volumetrically and qualitatively. The resolution of images being captured by the mobile cameras in smartphones, webcams or remote/in-office conference solutions is improving substantially. Which means more dense information is available at our disposal to break down, extract and utilize rich spatial information.

\begin{figure}[htbp]
\centerline{\includegraphics[width = 3.5in]{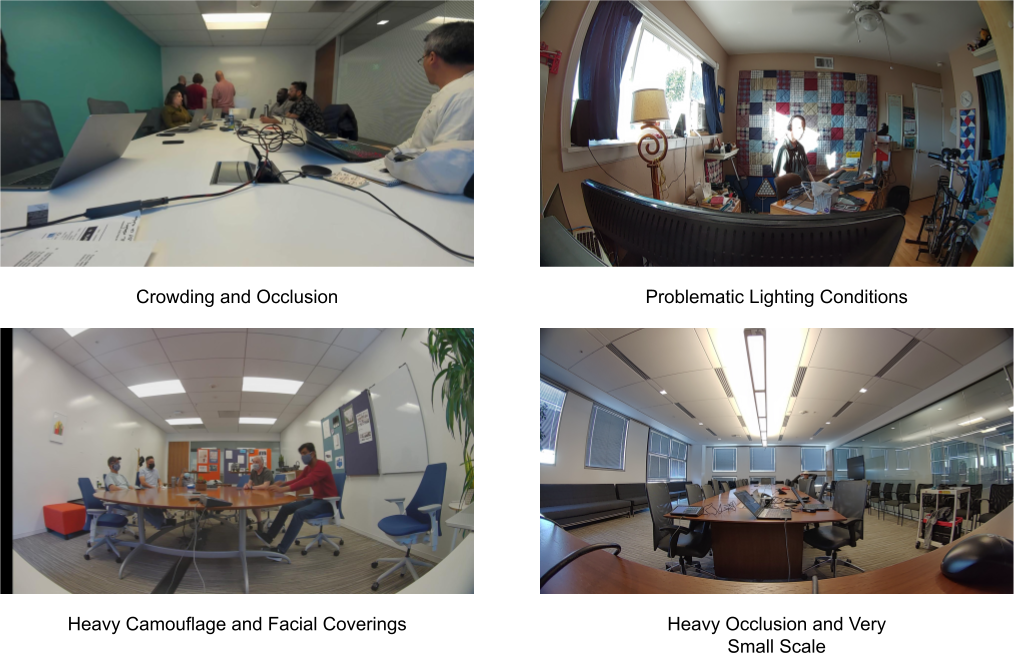}}
\caption{Challenging Scenarios}
\label{fig1}
\end{figure}

This paper describes our work in developing ASBU-net to further improve these on device high accuracy efficient neural network models. Some of the key challenges that we face in general and with most of these efficient models are detecting objects - in a crowded environments, partially/heavily occluded objects, significantly deformed objects, and most importantly the dynamic scale of objects. To address most of these problems it is very important to utilize the rich spatial and contextual information available in the image. There has also been a lot of work on detecting small objects, crowded environments and combination of both, but most of them are computationally extensive. Some of these challenges can be illustrated with the example images in figure below Fig \ref{fig1}.

Atrous/dilated convolutions \cite{b39} have successfully demonstrated capturing the spatial and contextual information, and simultaneously also reducing the computational complexity of deep CNNs with wider receptive fields. The popularity of atrous convolutions has led to its adoption in various efficient/mobile existing architectures like atrous spatial pyramid pooling, efficient spatial pyramid, etc. In our approach we utilize the benefits of using atrous convolutions too, but instead of applying aggressive parallel dilated convolution which might cause a loss in spatial information or aggressive series of dilated convolution, we propose a method which combines a parallel and series combination of atrous convolutional layers in network-in-network blocks.  

Our goal here is to have a model that is optimizes the accuracy-latency trade off on mobile devices and is easily deployable using most of standard frameworks. We tackle all the above mentioned problems along with keeping the model light by introducing the atrous space bender layers, which constitutes a major building block of the atrous space bender network. And then we use the technique from U-Net architecture \cite{b37} which seeks to retain high spatial frequency information by directly adding skip connections between early and late layers. The paper walks through the steps that were taken to solve each problem and experiments that demonstrate the efficacy and value of each technique evaluated with apple to apple comparison.

The paper is organized as follows. After the introduction of the topic and problem statement we discuss the related work and some overlap in Section 2. Later, in Section 3 we briefly introduce each concept and building blocks for the model construction. Section 4 reviews the experimentation results and benchmark results. Section 5 contains conclusions and future work.

\section{Related Work}

\subsection{Contextual Models}
Contextual information is important not just to “guess” small/blurry objects based on a prior, but to make sense of the world i.e. resolve ambiguity, and notice "unusual" things. Current context based networks show that global features or contextual interactions \cite{b41}\cite{b42}\cite{b43}\cite{b44}\cite{b45}\cite{b46} are beneficial in correctly classifying pixels for semantic segmentation. When training models there has been an increase in the use of transfer learning that was originally designed for classification networks. But the issue with this is that the details in the downsampled features map are lost, which has been addressed by learning the upsampling filters i.e. use of fractionally strided convolutions \cite{b40}. Another popular work adopts dense CRF which takes the FCN output and refines the segmentation boundaries. One of the other popular techniques that has been used widely is atrous/dilated convolution based, which has shown to preserve large receptive fields and give dense predictions. However, some of these strategies cause drop in efficiency, for example, PSPNet (Zhao et al.) \cite{b48} uses convolution on flat feature maps after pyramid pooling and upsampling. The other popular technique, DeepLab \cite{b39} uses the ASPP module that uses a parallel combination of dilated convolutions which causes two problems. First, it might cause failure in aggregating the local features due to sparsity of the kernels, and that is detrimental to small objects. Second, aggressive dilation rate also reduces to 1x1 convolution in extreme cases. The Atrous Space Bender-Net mitigates this by keeping a close to linear increase in receptive field rather than exponential.

\subsection{Skip Connections}
Skip connection has gained a lot of traction since the success of ResNet architecture. However, the concept of skip connections has been around prior to residual networks, for example, Highway Networks (Srivastava et al.) had skip connections with gates that controlled and learned the flow of information to deeper layers. Since then a lot of variations of skip connections have been designed like skip connections via addition in ResNet by He et al. in 2015 to solve the image classification problem, dense skip connections via concatenation by Huang et al. in 2017 and one of the most popular use cases long skip connections introduced in U-Net by Ronneberger et al. for biomedical image segmentation. The reason why skip connections are so popular because they alleviate the vanishing-gradient problem, strengthen feature propagation, encourage feature reuse, and substantially reduce the number of parameters. We leverage the long skip connections like a partial U-Net architecture to retain high spatial frequency information by directly adding skip connections between early and late layers. We dive into more details in the section \ref{ABSU_REF_MARKER}

\subsection{Network Compression}
In the recent years, there has been a slight shift in the paradigm where the focus moved towards having networks that can run on actual hardware with limited memory and compute capabilities \cite{b9}\cite{b10}\cite{b11}\cite{b12}\cite{b13}\cite{b14}\cite{b15}. Several approaches around network compression and small architectures have been researched and reported. However, there has always been a trade off between accuracy, size, latency and covering edge cases. This paper proposes an approach that has a configurable network to match the resource restriction, latency and size, still collecting contextual information and giving relatively better overall performance than existing architectures. 
 
SqueezeNet \cite{b9} uses a bottle neck design approach with extensive 1x1 convolutions with squeeze and expand layers. MobileNets, \cite{b14}\cite{b15}\cite{b29} which are built primarily from depthwise separable convolutions, improve upon the number of operations using inverted residuals and linear bottlenecks. ShuffleNet \cite{b30} utilizes group convolution and channel shuffle operations. MnasNet \cite{b31} introduces lightweight attention modules based on squeeze and excitation into the bottleneck structure. ShiftNet \cite{b32} proposes the shift operation interleaved with point-wise convolutions to replace expensive spatial convolutions. The other popular ways of smaller networks are based on quantization (compressing), shrinking and factorizing \cite{b33}\cite{b34}\cite{b35}. Denton et al. apply singular value decomposition (SVD) to a pretrained CNN model and Han et al. developed network pruning. We also use 8-bit integer quantization to further reduce the size, which is optional when a 32 bit float version is used or a full precision model is run.

\section{Model Architecture And it's Building Blocks}
In this section we will go over the basic building blocks that are used to make the ASBU-Net/LogiNet, which are atrous/dilated convolution, Atrous Space Bender Layer, and Atrous Space Bender Network. We later combined all the building blocks to create the final model design for the ASBU-Net. We will see how effective use of atrous/dilated convolution helps in collecting dense spatial features and thereby context around the objects. We will also discuss how even though we use the concept of feature pyramid network to extract smaller objects but still exploit the pyramidal hierarchy to construct feature pyramids with marginal extra cost. And we also go over the reasons behind design considerations.

\begin{figure*}[!tp]  
  \centering
  \includegraphics[scale=0.8]{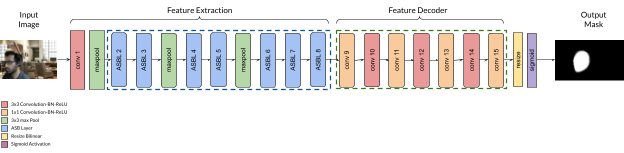}
  \caption{ASB-Net}
  \label{fig2}
\end{figure*}

\subsection{Atrous/Dilated Convolution}
Before we jump into describing the model modules, we will briefly discuss the concept of atrous convolution which we leverage to extract contextual information about the object and its surroundings. The core idea of atrous convolution is to increase the receptive field of different scales by changing the dilation rate while maintaining the size of the convolution kernel and obtain dense feature maps. It injects holes into the standard convolution map to expand the receptive field without increasing the parameters and channels. A 1D atrous convolution can be defined as follows: 

\begin{equation}
y[i]=\sum_{k=1}^{K} x[i+r.k]w[k]\label{eq1}
\end{equation}

where $y[i]$ is output signal, $x[i]$ is input signal with a filter $w[k]$ of length $K$, $r$ corresponds to the dilation rate to sample $x[i]$, and standard convolution is a special case for the rate $r=1$.

We will discuss in further detail as to how using atrous convolution with proper dilation rate plays an important role in avoiding some known problems and how we mitigate it by managing the receptive field growth close to linear rather than exponential.

\subsection{Atrous Space Bender Layer (ASBL)}
\begin{figure}[h]
\centerline{\includegraphics[width = 2.5in]{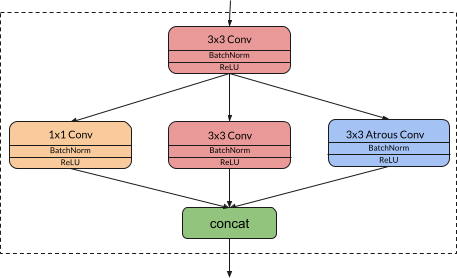}}

\caption{Atrous Space Bender Layer}
\label{fig3}
\end{figure}

The Atrous Space Bender layer forms the basic building block of the network. It consists of a squeeze 3x3 convolution layer connected to a parallel combination of 1x1 convolutions, 3x3 convolutions and 3x3 atrous convolutions, the output of these layers are concatenated and then fed to the next layer. We use a bottleneck like architecture with squeeze and expand layers to accumulate the features.

The 3x3 convolution layer helps squeeze the feature maps channelwise while still retaining the spatial information which is important to retain the contextual information. This is then followed by a parallel combination of - 
\begin{itemize}
\item pointwise convolution that looks at 1 pixel at a time which mimics fully connected behavior along the channel dimension and thereby helps retain relationship amongst channels
\item 3x3 convolution which as described earlier helps capture the spatial information from the immediate neighbors i.e. it captures more local information
\item 3x3 atrous convolution with increasing and then decreasing dilation rates which helps to increase the receptive field size, avoid downsampling, and generate a multiscale framework for segmentation i.e. it acts as a good trade-off between accurate localization (small field-of-view) and context assimilation (large field-of-view)
\end{itemize}

The hyperparameters or the tunable parameters for this layer are the number of filters in squeeze layer for ss\textsubscript{3x3} (spatial squeeze), e\textsubscript{1x1} (expand), se\textsubscript{3x3} (spatial expand), ase\textsubscript{3x3} (atrous spatial expand) and dilation rate for ase\textsubscript{3x3}. The number of filters in e\textsubscript{1x1} are set to be less than $(e_1x1 + se_3x3 + ase_3x3)$ to limit the number of channels to the expand layer. We maintain the squeeze:expand filter ratio below 1. 

\begin{figure*}[!tp]  
  \centering
  \includegraphics[scale=0.8]{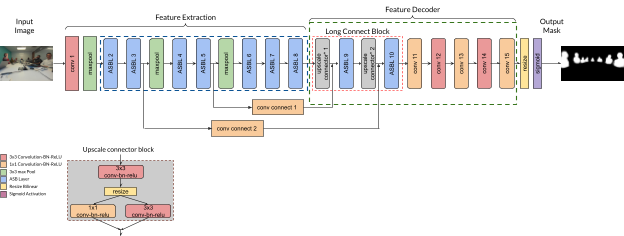}
  \caption{ASBU-Net}
  \label{fig4}
\end{figure*}

\subsection{Atrous Space Bender Network (ASB-Net)}
In this section we go over the architecture design and strategies for the entire ASB-Net which forms the backbone architecture to the ASBU-Net/LogiNet. As shown in the figure \ref{fig2} the ASB-Net begins with a primary/standalone (conv1) 3x3 convolution layer, followed by 7 ASB layers along with a couple of max pooling operations inserted in between to scale down the feature maps. The max pooling operations are performed to half the scale from the higher layer and we reduce the scale until $\frac{1}{16}^{th}$ of the input image dimension. There is also an option to reduce the scale until $\frac{1}{8}^{th}$ for a smaller model. The max pooling operations with a stride of 2 are placed after conv1, ASBL3, and ASBL5 for a network with $\frac{1}{16}^{th}$ scaling, and after conv1 and ASBL3 for a network with $\frac{1}{8}^{th}$ scaling. We minimize the use of max pooling operation to retain the information and also downsample late in the network to retain maximum amount of information using the strategy from SqueezeNet \cite{b9}. This network forms the encoder part of the network which learns the feature representations by use of convolutional and downsampling operations.

We gradually increase the number of filters per ASB-layer from the beginning to the end of the network. Another important aspect of this network is the selection of the dilation rates for each subsequent ASB-layer. Atrous layers when applied aggressively causes the problem where the local spatial information may be lost in the deeper layers and spatial inconsistency between neighboring units. Which means the purpose of using the atrous convolution which is to extract dense contextual information and decrease blurring in the segmentation masks is defeated. There are two solutions that we propose to combat this problem -
\begin{itemize}
\item[a.] increase the field of view of the ASB-layers almost linearly rather than exponentially
\item[b.] use a 3x3 and 1x1 convolution in parallel to atrous convolution, thereby retaining the local spatial information
\end{itemize}

To achieve an almost linear increase in receptive field, we deploy a strategy where we gradually increase the dilation/atrous rates in the ASB-layers and then decrease it from top to bottom. As we know the atrous convolution introduces holes in a convolutional kernel, while still maintaining the weights in the kernel. When the dilation rate for a kernel is set to some value let's say $\alpha$ that means we introduce a stride of $\alpha$ between each sample, thereby the receptive field of the kernel ($k>0$) is increased to $(\alpha(k-1) + 1)$ \cite{b36}. From this we can compute the receptive field, for the dilation rate of 1 the atrous conv behaves like a regular convolution operation i.e. the receptive field being equal to the size of the filter. Similarly, for dilation rates for 2,3,4, and 5 the receptive field for a 3x3 atrous convolution can be computed as - $2(3-1)+1 = 5$, 7, 9, and 11 respectively. Now when these atrous blocks are placed in series combination, the effective receptive field grows exponentially. The effective receptive field at each layer can be calculated using the formula from - Computing Receptive Fields of Convolutional Neural Networks \cite{b36}.

\begin{equation}
    r_0 = \sum_{l=1}^{L} \left((k_l-1)\prod_{i=1}^{l-1} s_i\right) + 1\label{eq2} \\
\end{equation}
where, \\
$k_l$: kernel size (positive integer)\\
$s_l$: stride (positive integer) \\
effective stride $S_l$ = $\prod_{i=l+1}^{L}s_i$ : the stride between a given feature map $f_l$ 		and the output feature map $f_L$

For the atrous convolution we use the above equation by replacing kernel size $k$ by $(\alpha(k-1) + 1)$. From the model diagram, dilation rates we use to define the local receptive field of ASB-layer keeps the effective receptive field in check. When we calculate the effective receptive field for each layer we see that the layerwise growth is almost linear and the flow of information of neighboring units is connected again. Hence, we show that our approach gradually recovers consistency between neighboring pixels and extracts local structure in higher layer.

\subsection{Atrous Space Bender U-Net/LogiNet}
\label{ABSU_REF_MARKER}

The ASBU-Net as the name suggests utilises the above backbone to not only improve the results but also to solve the problem of detecting objects at scale, by improving the contextual information gathered by atrous convolution blocks. As seen in figure \ref{fig4}, ASBU-Net is broken down into 2 major blocks - feature extraction block, and feature decoder block. The feature decoder block is further divided into 2 blocks - long connect block, which is a combination of upscale connector blocks and ASB layers, and decoder block.

A couple of important things that are different from traditional U-Net \cite{b37}, are the 1x1 convolution layer introduced in the skip connection and long connect blocks to which the skip connections feed into. The 3 reasons why we do this are - 
\begin{itemize}
\item[a.] To reduce the memory footprint from earlier layers
\item[b.] To capture additional spatial information and reduce semantic gap in the information
\item[c.] To retain information along the channels thereby reducing the gap from the encoding layers 
\end{itemize}

The number of filters from the higher level layers in the encoder block if passed directly to the decoder layer increases the memory footprint quite significantly. In order to keep the memory requirement to a minimum without losing information, we introduce 1x1 convolution to the skip connection to reduce the number of channels being passed down yet retaining the dense information. 

It is important to note that using the U-Net like architecture helps us to have feature reusability by connecting features from encoder to decoder. Thereby, helping recover information lost during the downsampling and encouraging the model to learn both the finer information (from layers close to input) and the more generic information (from layers close to output) of the objects. However, the shorter connections help in stabilizing the gradient updates. Significant problem with connecting the encoder directly with decoder is that, the layers in encoder contain low level feature information and as we go deeper into the network into the decoder block the feature information is supposed to be of much higher level. When these 2 layers are connected directly, there may be a semantic gap that gets introduced and thereby affecting the gradient propagation in the backward pass, which thereby also affects the accuracy. Inorder to solve this, we introduce what we call long connect blocks. For each skip connection a series combination of upscale connector block and ASB-layer with dilation rate of 1 (i.e. regular 3x3 convolution) is added to the long connect block. The gradual merging of the high level and low level features with upscale connector block and ASB-layer helps retain the information that gets lost in the sudden gap.

Post the long connect block, we add a combination of 1x1 convolution and 3x3 convolution, followed by bilinear resizing and sigmoid activation to obtain the semantic segmentation mask. Our goal while designing this network architecture was to keep it simple and use simplistic layers that can be easily quantized and run on embedded platforms.

\section{Experimental Setup and Hyperparameters}
In this section we discuss the experimental setup, training hyperparameters and validation metrics which we used to achieve competitive results compared to mobilenet-v2-unet and efficientnet-unet in quantitative evaluation on Logitech Validation dataset. We have also tried to provide a sufficient level of ablation study of the architecture and loss functions with qualitative results. Finally, we demonstrate that ASBU-net is an adaptable and extensible model for various use cases and can be easily deployed on almost any hardware without dramatic changes.

\subsection{Data and Augmentation}
We perform experiments on Logitech Video Collaboration proprietary segmentation dataset, containing 156k training images, which is split 80:20 during training for training and testing, and 4k validation images. Each image is annotated with segmentation polygons. We also use various data augmentation techniques to enhance the dataset during the runtime, like auto cropping, image flipping, adding gaussian noise, gamma variation, shadow augmentation, and color/saturation modification to name a few. For pre-processing operation we normalize the images before forward pass.

\subsection{Training Details}
Our entire training pipeline was implemented using Tensorflow backend \cite{b38}. The experiments were conducted on a desktop computer with Intel core i9-9900K processor (3.6 GHz, 16 MB cache) CPU, 64 GB RAM, and NVIDIA GeForce RTX 2080 Ti (11 GB, 1545 MHz) GPU.

For training the ASBU-net we used a momentum optimizer setting the momentum to 0.9, with a linear decay initialized to $10^{-2}$ , multiplying the decay rate by 0.1 at every 0.3 epoch. We use a very low value of $10^{-12}$ for l2 regularization. All convolution weights are initialized with Kaiming-He init \cite{b49}, and use batch-normalization layers with average decay of 0.99 for training higher batch sizes. As part of training,and test data, we have split the data-set into 80-20. The task of semantic segmentation is basically a problem of pixel wise classification, hence for the loss function we use weighted binary cross entropy. For convenience and ease of experimentation we have trained the network on a single class of segmenting humans in the frame.

\begin{figure*}[!tp]  
  \centering
  \includegraphics[scale=0.33]{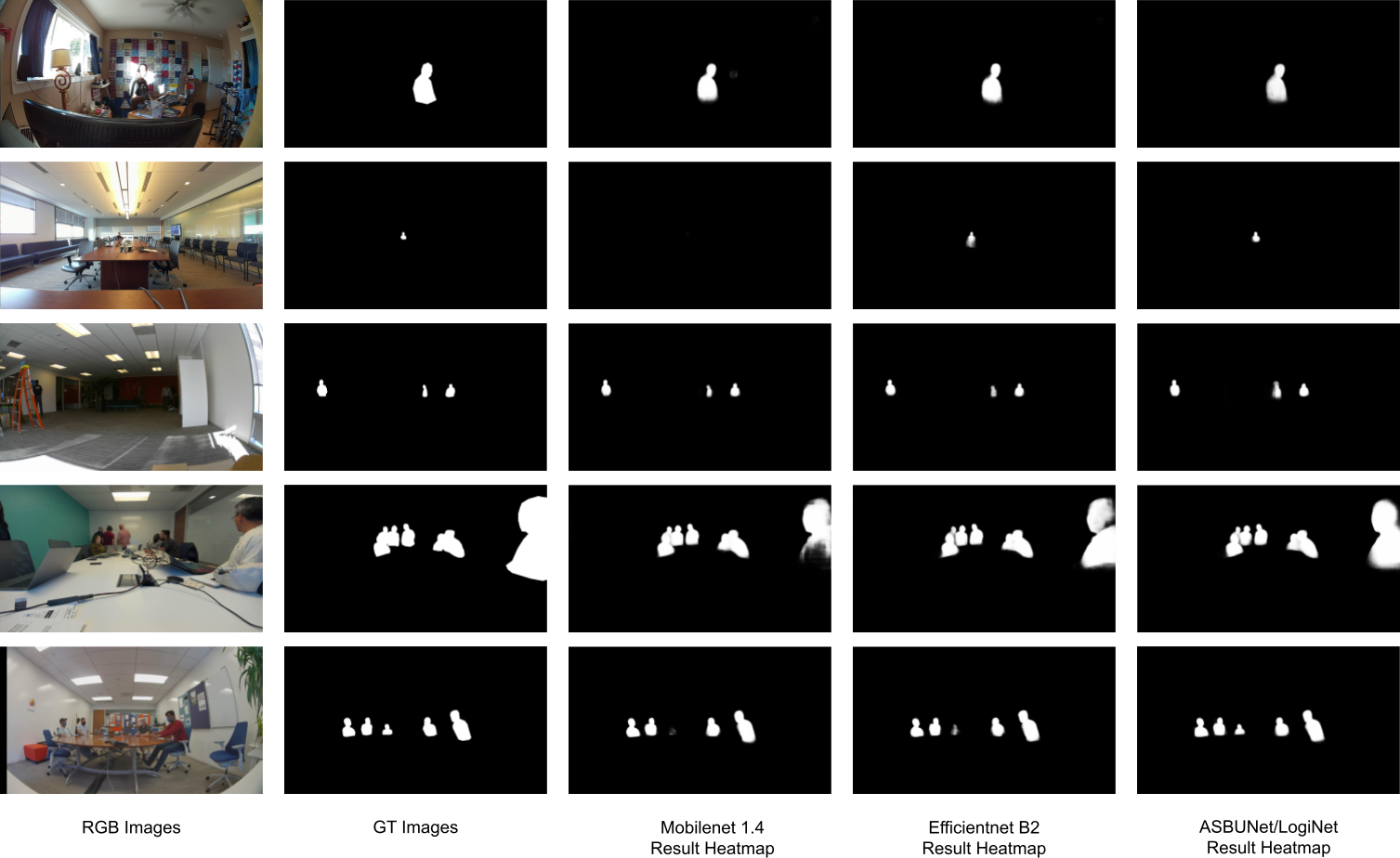}
  \caption{Results Heatmap Comparison}
  \label{fig5}
\end{figure*}

\subsection{Evaluation Metric}
It is important to decide a good evaluation metrics when validating the results of segmentation models. The metrics itself is not just enough. The labels need to be precise or closely encompassing the boundary of the actual object. Since most of the evaluation metrics consider IOU of the segmentation mask with the label mask, which means per pixel comparison, we usually end up penalizing the network for its performance around the edges if the labels are not accurate. Relative comparison between different models is not affected when the validation data remains same, but the overall score is what can be different. Inorder to combat this issue, we use ignore band which is calculated as inverse of bitwise XOR of dilated and eroded label around a label. This operation is performed based on the scale of the object to make sure size of object does not affect the original mask around smaller objects. Mathematically the ignore boundary band is given as follows -

\begin{equation}
\begin{aligned}
ll = Y \oplus J_{osf} \\
sl = Y \ominus J_{osf} \\
b_{ignore} = \neg(ll \veebar sl) \\
Y_{new} = Y \& b_{ignore} \\
Y_{pred\{new\}} = Y_{pred} \& b_{ignore}
\end{aligned}
\end{equation}

where,\\
osf = Objectwise scale factor \\
Y = original label mask \\
$Y_{pred}$ = predicted mask \\
ll = dilated label \\
sl = eroded label \\
$b_{ignore}$ = boundary to ignore \\

Once the new prediction map and label map are obtained using the above we then calculate the Jaccard index i.e. ratio of the intersection and union. In addition to this, we also add extra penalization for false detections by reducing the score by 1.0 for each misdetection.

\begin{figure}[!tp]
\centerline{\includegraphics[width = 3.5in]{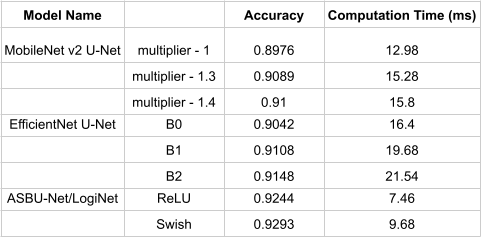}}
\caption{Results - Accuracy and Computation Time Comparison}
\label{fig6}
\end{figure}

\begin{figure}[!tp]
\centerline{\includegraphics[width = 3.5in]{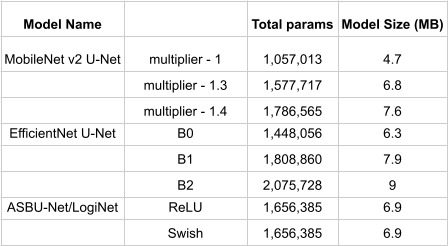}}
\caption{Results - Total Params and Model Size Comparison}
\label{fig7}
\end{figure}

\subsection{Results}
In this section, we go over the results from experimenting with ASBU-Net/LogiNet on the validation images from Logitech Video Collaboration dataset. A large-scale dataset containing high quality pixel-level annotations of 160,000 images. We use the 4000 images reserved for validation/test for obtaining the validation scores using the above explained evaluation metrics. It is also important to note that inorder to maintain the sanctity of the experiments we used the exact same setup for training and testing for all the 3 models which are ASBU-Net/LogiNet, Mobilenet-v2-Unet, and Efficient Net Unet.

Let's first look at the results subjectively on the heatmaps in the figure \ref{fig5}. We can see that most of these scenarios are pretty complex to detect humans with complex lighting conditions, crowding, heavy camouflage, and heavy occlusions. The subjective analysis shows that mobilenetv2 \cite{b15} based models have higher false positives and weak detections in uneven lighting conditions, with objects at different scales. The efficientnet\cite{b47} based models however, perform well with varied lighting conditions but have weak detections for objects that are closer and heavily occluded. Both mobilenetv2 \cite{b15} and efficientnet \cite{b47} based models fail to detect heavily camouflaged people of color at the end of the table. Under all these and more conditions we see that the ASBU-net/Loginet performs consistently well. 

As seen in the results table in figure \ref{fig6} and \ref{fig7}, we compare the ASBU-Net/LogiNet in terms of accuracy, total params, model size and computation time. We show that the ASBU-Net/LogiNet is pretty consistent in terms of performance in all these areas and still computationally inexpensive as compared to the other popular models. The validation accuracy is almost 3\% better than Mobilenet v2 U-Net with a multiplier 1.4, and almost 1.5\% better than EfficientNet U-Net version B2. We also show that the computation time is almost 3x faster for best performing efficient net and 1.5x$\sim$2x faster than mobilenet backend models. The computation times are on a workstation running Intel(R) Core(TM) i9-9900K CPU \@ 3.60GHz with NVIDIA GeForce RTX 2080 SUPER graphic card. ASBU-Net has 1.25x less parameters as compared to efficientnet \cite{b47} based B2 model, and $\sim$1.1x less parameters than mobilenetv2 \cite{b15} backed model, which means it will have lower power consumption and lower battery discharge on embedded platforms.

\section{Conclusion}
In this paper we present a novel model architecture based on dilated convolutions that utilizes the contextual information to perform precise image segmentation tasks. We also show that the general problems associated with the dilated convolutions can be handled by using the ASB-layer which is lightweight, demands less memory and computation, and can extract complex contextual feature representations. We show that the network architecture is more efficient, consistent and robust in comparison to some famous existing architectures. Furthermore, we also share our validation method which helps us ignore the labeling inefficiencies around the edges of the image and focus on the actual detections. We show that using the ASBU-Net you can detect objects that are at different scales, crowded, deformed, or partially occluded, and other complex conditions, and still maintain mobility and latency. We are certain that ASBU-net/Loginet can be used for a wide variety of tasks and more ablation studies can be performed to test more performance boost.

\pagebreak

\section{Supplementary}
\subsection{Model Layer Chart}
\begin{figure}[H]  
  \includegraphics[scale=0.5]{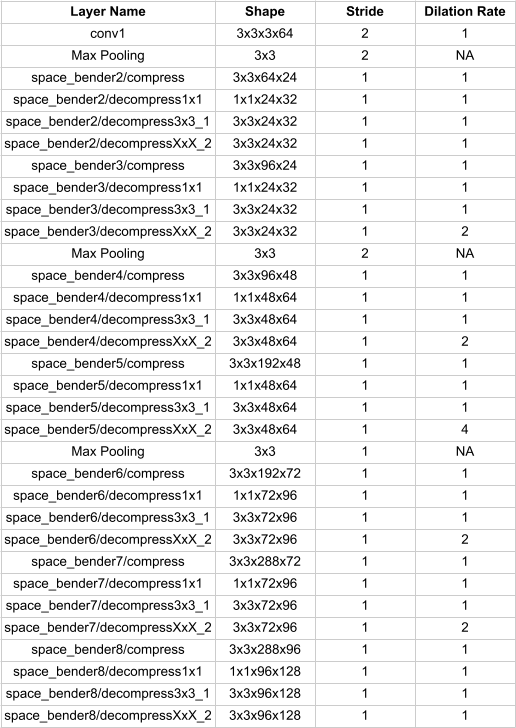}
  \caption{ASBU-Net Encoder Layer Specifications}
  \label{fig8}
\end{figure}

\end{document}